\documentclass[letterpaper]{article} 
\usepackage{aaai2027}
\usepackage[hyphens]{url}  
\usepackage{graphicx} 
\usepackage{natbib}  
\usepackage{caption} 
\usepackage{algorithm}
\usepackage{algorithmic}

\usepackage{newfloat}
\usepackage{listings}
\DeclareCaptionStyle{ruled}{labelfont=normalfont,labelsep=colon,strut=off} 
\floatstyle{ruled}
\newfloat{listing}{tb}{lst}{}
\floatname{listing}{Listing}

\usepackage{booktabs}
\usepackage{tabularx}
\usepackage{amsmath}
\usepackage{multirow}
\usepackage{makecell}
\usepackage{array}
\usepackage{xcolor}
\usepackage{colortbl}

\graphicspath{{figures/}}

\definecolor{splitgray}{RGB}{235,235,235}
\definecolor{ourblue}{RGB}{214,229,252}

\newcommand{\best}[1]{\textbf{#1}}
\newcommand{\second}[1]{\underline{#1}}

\newcommand{\textttt}[1]{\texttt{#1}}
\newcommand{\method}{\textbf{\textttt{LatentSkill}}}

\title{\method{}: From In-Context Textual Skills to In-Weight Latent Skills for LLM Agents}

\author{
  \textbf{Aofan Yu}$^{1}$\thanks{Equal contribution.} \quad
  \textbf{Chenyu Zhou}$^{1}$\footnotemark[1] \quad
  \textbf{Tianyi Xu}$^{1}$ \quad
  \textbf{Zihan Guo}$^{2,3}$ \quad
  \textbf{Rong Shan}$^{1}$ \\
  \textbf{Zhihui Fu}$^{4}$ \quad
  \textbf{Jun Wang}$^{4}$\thanks{Corresponding authors.} \quad
  \textbf{Weiwen Liu}$^{1}$\footnotemark[2] \quad
  \textbf{Yong Yu}$^{1}$ \quad
  \textbf{Weinan Zhang}$^{1,3}$\footnotemark[2] \quad
  \textbf{Jianghao Lin}$^{1}$\footnotemark[2]
}

\affiliations{
  \normalfont
  $^{1}$Shanghai Jiao Tong University \quad
  $^{2}$Sun Yat-Sen University \\
  $^{3}$Shanghai Innovation Institute \quad
  $^{4}$OPPO Research Institute \\[4pt]
  \texttt{junwang.lu@gmail.com} \quad
  \texttt{\{wwliu, wnzhang, linjianghao\}@sjtu.edu.cn}
}

\begin{document}
\nocopyright
\maketitle


\let\originalfootnote\footnote
\renewcommand{\footnote}[1]{}

\begin{abstract}
Agent systems increasingly use textual skills to encode reusable task procedures, but injecting these skills into the prompt at every step incurs substantial context overhead and exposes skill content as plaintext. We present \textbf{\method{}}, a framework that converts textual skills into plug-and-play LoRA adapters through a pretrained hypernetwork. \method{} stores skill knowledge in weight space rather than context space, removing per-step skill tokens while preserving modular loading, scaling, and composition. On ALFWorld and Search-QA, \method{} outperforms the corresponding in-context skill baseline while using substantially fewer prefill tokens: it improves ALFWorld success by 21.4 and 13.4 points on the seen and unseen splits with 63.9\% fewer prefill tokens on average, and improves Search-QA exact match by 3.0 points while using 71.8\% fewer tokens per step. Further analysis shows that generated skill LoRAs form a structured semantic geometry, can be continuously modulated via the LoRA scaling coefficient, and can be composed through parameter-space arithmetic when skill components are aligned. 
These findings suggest that weight-space skills provide an efficient, modular, and less exposed substrate for extending LLM agents.
\end{abstract}

\let\footnote\originalfootnote

\section{Introduction}
\begin{figure}[t]
    \centering
    \includegraphics[width=0.84\columnwidth]{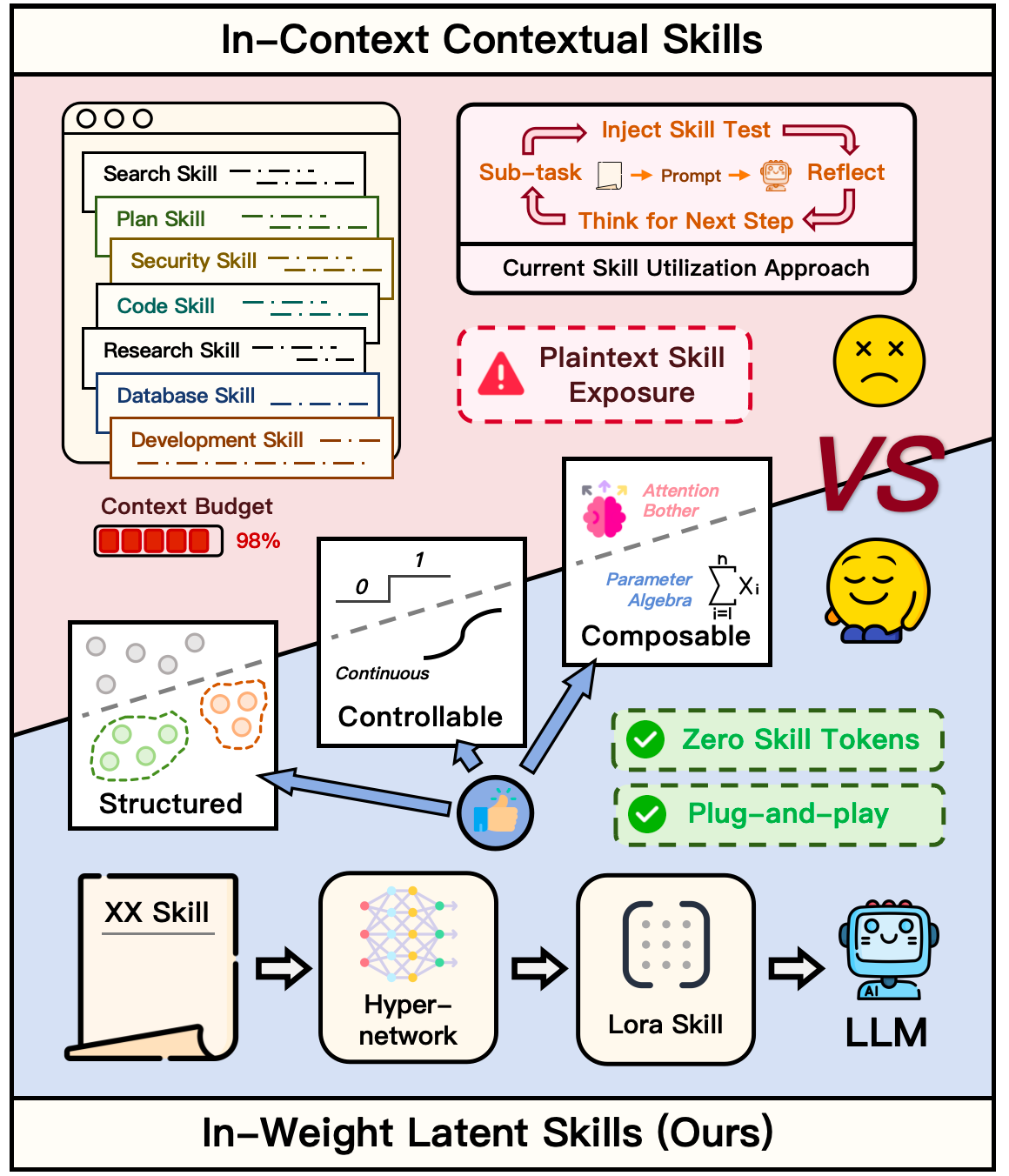}
    \caption{The key advantages of \method{} over in-context skill: (1) zero skill tokens in the prompt with plug-and-play modularity, and (2) a structured, controllable, and composable skill weight space.}
    \label{fig:motivation}
\end{figure}
LLM agents increasingly solve complex tasks by interleaving reasoning, action, and feedback from external environments \citep{Yao2023ReAct, Shinn2023Reflexion, Zhao2024ExpeL}.
To handle specialized and long-horizon tasks, many systems further rely on external skills: reusable textual procedures that encode task strategies, tool-use patterns, and recovery heuristics \citep{Wang2023Voyager, Xia2026SkillRL, Wu2025EvolveR, SkillOS2026, pan2026skill, wang2026skill}.
A common design retrieves relevant skills from a skill library and inserts them into the prompt when the agent selects an action \citep{Cho2026SkillRet, zhang2026mmskill}.
This design is simple and modular, but it becomes costly as interactions grow longer and skill libraries grow larger.
The same skill text may be inserted repeatedly across decision steps, consuming context and increasing prefill cost. Long inputs also make it harder for models to use all supplied information robustly \citep{Jiang2023LLMLingua, Jiang2024LongLLMLingua, Liu2024LostMiddle}.
Moreover, skills kept as readable prompt content can expose proprietary procedures and share the instruction channel with potentially untrusted observations \citep{Greshake2023IndirectInjection, Liu2023PromptInjection, Wallace2024InstructionHierarchy, Schmotz2026SkillSecurity, guo2026skill}.
Parametric alternatives, such as agent fine-tuning or curriculum learning, avoid inserting skill text at inference time, but they fuse skills into the model parameters \citep{Chen2023FireAct, Zeng2024AgentTuning, Lu2026Skill0}.
As a result, individual skills become difficult to update, remove, or combine.
Existing approaches therefore face a trilemma: how to avoid repeated skill text in the prompt, while still allowing skills to be updated modularly and combined during inference.


We introduce \textbf{\method{}}, a framework that converts textual agent skills into LoRA adapters through hypernetwork-based adapter generation \citep{Ha2017Hypernetwork, Hu2022LoRA, Liu2026SHINE}.
Instead of delivering skills through the context window, \method{} represents them in weight space.
Given a skill description, a trained hypernetwork generates a skill-specific LoRA adapter in a single forward pass, which is then mounted on the backbone LLM during inference.
The original skill text is no longer included in the prompt, reducing both context cost and exposure as readable text.
Meanwhile, the generated adapters remain modular: they can be loaded, unloaded, replaced, or scaled without retraining the backbone, aligning with prior work on non-destructive adapter composition and dynamic LoRA combination \citep{Pfeiffer2021AdapterFusion, Huang2024LoraHub}.
They can also be combined when the original skills are decomposed into semantically aligned components.

Beyond efficiency, we show that representing skills as LoRA weights gives them useful structure.
Across ALFWorld \citep{Shridhar2021ALFWorld} and Search-QA \citep{Jin2025SearchR1}, \method{} improves over the direct in-context skill baseline while using the same skill descriptions and substantially reducing the prompt overhead caused by skill text.
We further find that generated skill LoRAs are \textbf{structured}, with skills from different domains forming separable clusters in weight space; \textbf{controllable}, since their effect can be adjusted through the LoRA scaling coefficient $\alpha$; and \textbf{composable}, when skill descriptions are decomposed into aligned components before their LoRAs are combined.
Together, these results suggest that latent skill weights are not just compressed prompts, but a representation of procedural knowledge that can be inspected, adjusted, and reused.

Our contributions are fourfold.
\textbf{(1)} We develop a two-stage, multi-domain training pipeline for agent skills: document reconstruction and completion pretraining on approximately 171K publicly available skill documents spanning diverse domains, followed by trajectory-supervised fine-tuning across multiple agent tasks.
\textbf{(2)} We instantiate generated adapters as token-free, plug-and-play \method{} modules that can be cached and mounted without retaining the original skill text, substantially reducing repeated inference-time context overhead.
\textbf{(3)} We show that \method{} consistently outperforms direct in-context skill prompting on ALFWorld and Search-QA while eliminating the context overhead introduced by skill text.
\textbf{(4)} We systematically characterize the resulting \method{} space, revealing domain-level structure, continuous control through injection scaling, and parameter-space composition when skills are decomposed into semantically aligned components.


\section{Related Work}
\subsection{LLM Agents and Skill Systems}
 
LLM agents solve complex tasks by interleaving reasoning and action \citep{Yao2023ReAct} and can improve from failures through self-reflection \citep{Shinn2023Reflexion}. To extend agents beyond the knowledge boundary of a single model, a growing body of work has explored injecting external experiential knowledge at decision time \citep{zhou2026externalization}. Early approaches store raw trajectories or reflective summaries in external memory banks and retrieve them into the context during inference \citep{Zhao2024ExpeL, Chhikara2025Mem0,zhang2026mmskillsmultimodalskillsgeneral}. Subsequent work converged on skills as the core abstraction: \citet{Wang2023Voyager} introduced an ever-growing library of executable skills for embodied agents; \citet{Xia2026SkillRL} distill trajectories into a hierarchical SkillBank that co-evolves with the agent's policy through recursive reinforcement learning; and \citet{Wu2025EvolveR} refine interaction experience into reusable strategic principles.  On the industry side, Anthropic's Agent Skills specification has been adopted by mainstream harnesses including Claude Code, Cursor, and Gemini CLI. All of the above deliver skills as natural-language text injected into the context window. \citet{Lu2026Skill0} take a different direction by internalizing skills into model parameters through a training-time curriculum that progressively withdraws skill context, enabling zero-shot execution at inference. Their skills, however, are irreversibly fused into the backbone, forfeiting modular flexibility. In contrast, \method{} encodes skills as modular LoRA adapters that remain separate from the frozen backbone parameters, eliminating the need to include skill text in the backbone’s inference prompt while retaining plug-and-play flexibility.
 
\subsection{Hypernetworks for LoRA Generation}
 
Hypernetworks generate the weights of a target network through a separate learned network \citep{Ha2017Hypernetwork} and have recently been applied to produce LoRA adapters \citep{Hu2022LoRA} for LLMs in a single forward pass, bypassing the iterative overhead of per-task fine-tuning. 
Text-to-LoRA \citep{charakorn2025texttolorainstanttransformeradaption} uses a shared MLP hypernetwork conditioned on task, module, and layer representations to generate LoRA weights, but does not explicitly model interactions among target layers during parameter generation. 
Generative Adapter \citep{Chen2025GenAdapter} utilizes the hidden states of the backbone LLM to produce LoRA weights, yet its layer-specific generators make module coverage costly in the reported implementation. 
ICAE \citep{ge2024incontextautoencodercontextcompression} compresses contexts into a fixed set of compact tokens that directly condition subsequent generation, rather than generating LoRA weights, and is constrained by the resulting information bottleneck. 
SHINE \citep{Liu2026SHINE} extracts memory states at every layer and enables bidirectional cross-layer information flow through an attention mechanism, but does not analyze the intrinsic structure of the generated LoRA weight space. 
Doc-to-LoRA \citep{charakorn2026doctoloralearninginstantlyinternalize} generates adapters through a Perceiver-style encoder, but likewise focuses on context internalization. Skill-to-LoRA \citep{Zhang2026SkillToLoRA} trains an independent adapter for each skill using skill-conditioned synthetic demonstrations, whereas Parametric Skills \citep{Zhao2026ParametricSkills} trains a hypernetwork to map free-form skills into LoRA adapters for software-engineering tasks. 
In comparison, \method{} targets the general agent skill encoding setting and provides a unified characterization of the semantic geometry, scaling-based behavioral control, and alignment-dependent composition of hypernetwork-generated skill LoRAs.
\section{Method}
\label{sec:method}
\begin{figure*}[t]
    \centering
    \includegraphics[width=\textwidth]{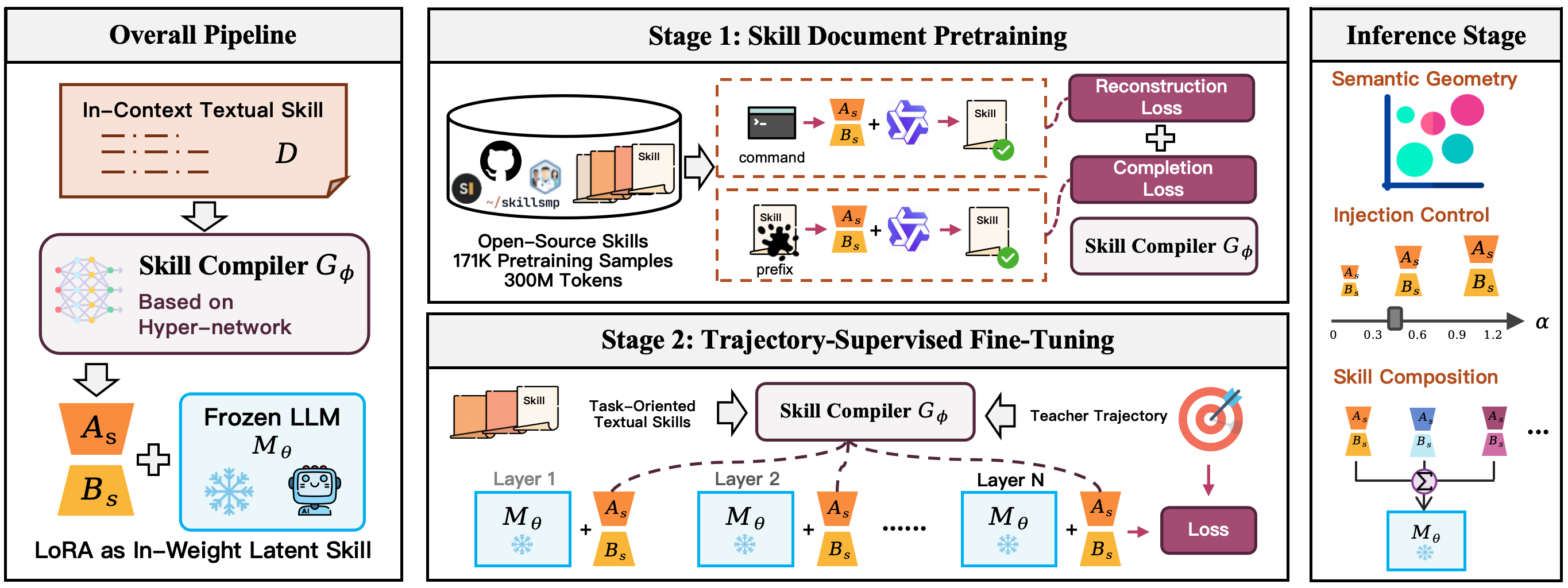}
    \caption{
    Overview of \method{}.
    \textbf{Left}: textual skills are transformed into in-weight latent skills through hypernetwork-based LoRA generation.
    \textbf{Middle}: the skill compiler is trained by skill document pretraining and trajectory-supervised fine-tuning.
    \textbf{Right}: the resulting latent skills support structured semantic geometry, controllable injection strength, and composable parameter-space arithmetic at inference time.
    }
    \label{fig:method}
\end{figure*}

\method{} converts a textual skill into a LoRA adapter that can be mounted on a frozen LLM. The skill text is consumed by a skill compiler, while inference uses the generated adapter instead of inserting the skill document into the prompt. We describe it in four parts: latent skill definition, document-level pretraining, trajectory-supervised fine-tuning, and inference-time control and composition.

\subsection{Latent Skill Definition}
\label{sec:latent_skill_definition}

Let $M_\theta$ denote a backbone LLM with frozen parameters $\theta$, and let $s$ be a textual skill document. At decision step $t$, the agent observes a history $h_t$ and produces an output $y_t$. In-context skill prompting directly conditions the model on the skill text. \method{} instead replaces textual conditioning with parameter conditioning.

A skill compiler $G_\phi$ maps the skill document to a set of LoRA updates:
\begin{equation}
    \Delta_s = G_\phi(s),
\end{equation}
where $\Delta_s$ denotes the latent skill induced by $s$. With $\Delta_s$ mounted, the model predicts from the task history alone:
\begin{equation}
    p_{\theta,\phi}(y_t \mid h_t, s)
    =
    p_{\theta \oplus \alpha \Delta_s}(y_t \mid h_t),
\end{equation}
where $\oplus$ denotes LoRA-based parameter augmentation and $\alpha$ controls the injection strength.

For each target module $m \in \mathcal{M}$, the generated update follows the standard LoRA form $\Delta W_s^{(m)}=B_s^{(m)}A_s^{(m)}$. Mounting a latent skill adds a scaled low-rank update to the frozen weight:
\begin{equation}
    W'
    =
    W
    +
    \frac{\alpha}{r} B_s A_s ,
\end{equation}
where $r$ is the LoRA rank. A complete latent skill is the collection of generated updates $\Delta_s=\{\Delta W_s^{(m)}\}_{m\in\mathcal{M}}$ over the selected target modules. 

\subsection{Skill Document Pretraining}
\label{sec:pretraining}

First, we pretrain the skill compiler on a corpus
of textual skill documents, denoted by \(\mathcal{D}_{pre} =
\{s_i\}_{i=1}^N\). The goal is to initialize \(G_\phi\) to map procedural text into usable adapter weights while keeping the backbone LLM frozen.
Given a skill document \(s\), we sample the reconstruction and completion objectives with equal probability, resulting in a 1:1 mixture of the two document-level pretraining tasks. In the reconstruction task, the compiler reads the complete skill document \(s\), and the adapted backbone receives a reconstruction instruction as input and is trained to reproduce the original document \(s\). In the completion task, we construct a truncated prefix \(\tilde{s}\) by randomly removing the latter part of the document; the compiler reads \(\tilde{s}\), and the adapted backbone is trained to complete the full skill document.

For each skill, we construct document-level supervision instances $(s_i^{\mathrm{src}}, q_i, z_i)$, where $s_i^{\mathrm{src}}$ is the text provided to the compiler, $q_i$ is the prompt given to the adapted backbone, and $z_i$ is the target output. Let $\Delta_i^{\mathrm{pre}}=G_\phi(s_i^{\mathrm{src}})$. The pretraining objective is
\begin{equation}
    \mathcal{L}_{\mathrm{pre}}
    =
    -\sum_{i,j}
    \log p_{\theta \oplus \alpha \Delta_i^{\mathrm{pre}}}
    \bigl(z_{i,j} \mid q_i, z_{i,<j}\bigr),
\end{equation}
where the summation ranges over all document-level supervision instances and target tokens.

Only the compiler parameters $\phi$ are updated. Since the skill document is provided to $G_\phi$ rather than directly to the adapted backbone, information useful for predicting $z_i$ must be mediated through the generated adapter.

\subsection{Trajectory-Supervised Fine-Tuning}
\label{sec:sft}

After pretraining, we fine-tune the skill compiler with teacher agent trajectories. Let $\mathcal{D}_{\mathrm{sft}}=\{(s_i,\tau_i)\}_{i=1}^{M}$ denote the supervised dataset. Each example pairs a skill document $s_i$ with a teacher trajectory $\tau_i=\{(h_{i,t}, y_{i,t}^{\star})\}_{t=1}^{T_i}$, where $h_{i,t}$ is the agent history at step $t$ and $y_{i,t}^{\star}$ is the teacher output.

For each pair $(s_i,\tau_i)$, the compiler generates one latent skill, denoted by $\Delta_i^{\mathrm{sft}}=G_\phi(s_i)$. The same adapter is mounted throughout the entire trajectory. The fine-tuning objective is
\begin{equation}
    \mathcal{L}_{\mathrm{sft}}
    =
    -\sum_{i,t,j}
    \log p_{\theta \oplus \alpha \Delta_i^{\mathrm{sft}}}
    \bigl(y_{i,t,j}^{\star}
    \mid h_{i,t}, y_{i,t,<j}^{\star}\bigr),
\end{equation}
where the summation ranges over all trajectories, decision steps, and target tokens.

The backbone remains frozen, and only $\phi$ is updated. Since $\Delta_i^{\mathrm{sft}}$ is generated solely from the skill document $s_i$ and shared across all decision steps in $\tau_i$, the objective encourages the adapter to capture skill-level, trajectory-consistent policy information rather than per-step adaptations. This aligns the compiler to produce latent skills whose effects remain stable across multi-step interaction.

\subsection{Inference-Time Skill Control and Composition}
\label{sec:inference}

At inference time, skill compilation is separated from agent execution. Given a skill library $\mathcal{S}=\{s_1,\ldots,s_K\}$, each skill can be compiled once and stored in an adapter cache $\mathcal{C}[k]=G_\phi(s_k)$. After compilation, the skill is not included in the prompt.

For a task instance, a skill selector chooses one or more relevant skills. If a single skill $s_k$ is selected, its cached adapter $\mathcal{C}[k]$ is mounted on the backbone with injection coefficient $\alpha_k$. The agent then predicts each step from the current history $h_t$ using the adapted model. Setting $\alpha_k=0$ recovers the frozen backbone, while larger values increase the influence of the latent skill.

When multiple skills are selected, \method{} composes their adapters in weight space:
\begin{equation}
    \Delta_{\mathcal{K}}
    =
    \sum_{k\in \mathcal{K}}
    \alpha_k \mathcal{C}[k],
\end{equation}
where $\mathcal{K}$ is the selected skill set. The composed adapter is then mounted on the LLM for inference.

For skills with shared subcomponents, direct adapter addition may over-amplify common behavior. \method{} therefore also supports component-level composition. Specifically, a skill can be decomposed into semantic components $s_k=\{c_{k,1},\ldots,c_{k,L_k}\}$, and each component can be compiled independently as $\Delta_{k,\ell}=G_\phi(c_{k,\ell})$, and the final adapter can be formed by adding retained shared and skill-specific components, e.g., $\Delta_{\mathrm{comp}}=\sum_{c\in\mathcal{U}}\gamma_cG_\phi(c)$, where $\mathcal{U}$ is the selected component set and $\gamma_c$ is an optional component-level injection coefficient.

\section{Experiments}

\subsection{Experiment Setup}
\paragraph{Benchmarks.}
We evaluate \method{} on two agent benchmarks. \textbf{ALFWorld} \citep{Shridhar2021ALFWorld} is a text-based interactive environment aligned with the ALFRED embodied AI benchmark, comprising six categories of tasks: Pick and Place (Pick), Look at Obj in Light (Look), Pick Clean then Place in Recep (Clean), Pick Heat then Place in Recep (Heat), Pick Cool then Place in Recep (Cool), and Pick Two Obj and Place (Pick2). \textbf{Search-QA} follows the evaluation protocol of \citet{Jin2025SearchR1} and covers seven QA datasets, including three single-hop benchmarks (NQ \citep{Kwiatkowski2019NQ}, TriviaQA \citep{Joshi2017TriviaQA}, PopQA \citep{Mallen2023PopQA}) and four multi-hop benchmarks (HotpotQA \citep{Yang2018HotpotQA}, 2WikiMultihopQA \citep{Ho2020Wiki}, MuSiQue \citep{Trivedi2022MuSiQue}, Bamboogle \citep{Press2023Bamboogle}). Training data are drawn from NQ and HotpotQA, and the remaining five datasets serve as out-of-domain evaluation sets.
 

\paragraph{Baselines.}
We organize the baselines into three categories.
\textit{Prompt} baselines include Vanilla, Few-shot, CoT \citep{Wei2022CoT}, Reflexion \citep{Shinn2023Reflexion}, AdaPlanner \citep{Sun2023AdaPlanner}, R1-Instruct, RAG \citep{Lewis2020RAG}, and In-context Skill. In-context Skill uses the same skill as \method{} but places it in the prompt.
\textit{Fine-tuning} baselines include Full SFT, Shared LoRA, and Per-skill LoRA. Full SFT and Shared LoRA use the same teacher trajectories as \method{} to train all model parameters and a single adapter shared across skills, respectively. Both retain the skill in the prompt at inference, controlling for the effect of teacher-trajectory supervision. Per-skill LoRA trains a separate adapter using the teacher trajectories associated with each skill and directly loads the corresponding adapter at inference without a skill prompt. 
\textit{Hypernetwork} baselines include SHINE \citep{Liu2026SHINE} and Text-to-LoRA \citep{charakorn2025texttolorainstanttransformeradaption}, both trained on the same teacher trajectories as \method{}, enabling a controlled comparison with other hypernetwork-based adapter generation methods.

 
\paragraph{Implementation Details.}
We use Qwen3-8B as the frozen backbone LLM. The compiler follows the same architecture as the one in SHINE \citep{Liu2026SHINE}. It generates rank-8 LoRA adapters for the query, key,
value, output, gate, up, and down projection modules in every Transformer layer. In the pretraining stage, we train the compiler on approximately 171K deduplicated skill documents crawled from GitHub, totaling roughly 300M tokens. In the SFT stage, we use the skill library and teacher trajectories released by \citet{Xia2026SkillRL}. The trajectories were generated by OpenAI o3 and comprise 237 complete ALFWorld task trajectories and 500 complete Search-QA task trajectories. Each complete trajectory contains the full sequence of observations, reasoning traces, and agent actions from task initialization to termination. We mix them into a single training set covering 5 ALFWorld skills and 3 Search-QA skills. For Search-QA, passage retrieval is performed using E5~\citep{Wang2022E5}. Our evaluation isolates skill encoding and execution under correct skill selection. Both stages are trained for 10 epochs on 8$\times$H100 GPUs using AdamW with a weight decay of 0.1 and a maximum sequence length of 4,096. Pretraining uses a batch size of 64, a learning rate of $5\times10^{-5}$, and 200 warmup steps, while SFT uses a batch size of 32, a learning rate of $1\times10^{-5}$, and 400 warmup steps. 

\subsection{Main Results}

\newlength{\alfcategorywidth}
\newlength{\alfmethodwidth}
\setlength{\alfcategorywidth}{1.85cm}
\setlength{\alfmethodwidth}{2.20cm}

\begin{table*}[t]
\centering
\small
\setlength{\tabcolsep}{2pt}
\renewcommand{\arraystretch}{1.05}

\begin{tabularx}{\textwidth}{
  @{}
  >{\centering\arraybackslash}p{\alfcategorywidth}
  >{\raggedright\arraybackslash}p{\alfmethodwidth}
  *{10}{>{\centering\arraybackslash}X}
  @{}
}
\toprule
\multirow{2}{*}{\textbf{Category}}
  & \multirow{2}{*}{\textbf{Method}}
  & \multicolumn{6}{c}{\textbf{ALFWorld Task}}
  & \multirow{2}{*}{\textbf{Avg}$\uparrow$}
  & \multirow{2}{*}{\textbf{Step}$\downarrow$}
  & \multicolumn{2}{c}{\textbf{Cost}$\downarrow$} \\
\cmidrule(lr){3-8}
\cmidrule(lr){11-12}
  & & \textbf{Pick} & \textbf{Look} & \textbf{Clean}
  & \textbf{Heat} & \textbf{Cool} & \textbf{Pick2}
  & & & \textbf{Prefill} & \textbf{Decode} \\

\specialrule{\lightrulewidth}{\aboverulesep}{0pt}
\rowcolor{black!7}
\multicolumn{12}{c}{\textit{Seen Split}} \\
\specialrule{\lightrulewidth}{0pt}{\belowrulesep}

\multirow{5}{*}{\makecell[c]{Prompt}}
  & Vanilla
  & 82.9 & 46.2 & 18.5 & 37.5 & 32.0 & 29.2 & 43.6
  & 35.0 & \best{0.44} & 0.55 \\

  & Few-Shot
  & 82.9 & 46.2 & 44.4 & \best{68.8} & 36.0 & 12.5 & 50.0
  & 31.3 & 2.04 & 0.54 \\

  & Reflexion
  & 77.1 & 53.9 & 33.3 & 31.3 & \second{56.0} & 12.5 & 46.4
  & 33.3 & 0.55 & 0.66 \\

  & AdaPlanner
  & \second{91.4} & 30.8 & 33.3 & 37.5 & 28.0 & 33.3 & 47.1
  & 35.9 & 0.54 & 0.84 \\

  & In-Context Skill
  & 85.7 & \second{69.2} & \best{70.4} & 31.3 & 12.0 & 33.3
  & 52.9 & 30.8 & 1.21 & 0.50 \\
\midrule


\multirow{3}{*}{\makecell[c]{Fine-tuning}}
  & Full SFT
  & 82.9 & 38.5 & \best{70.4} & 43.8 & 24.0 & 37.5
  & 53.6 & 36.0 & 1.38 & 0.45 \\

  & Shared LoRA
  & 68.6 & 53.9 & 59.3 & 31.3 & 16.0 & 29.2
  & 45.0 & 38.9 & 1.34 & 0.47 \\

  & Per-skill LoRA
  & 71.4 & 46.2 & 29.6 & \second{56.3} & 36.0 & 29.2
  & 45.7 & 38.5 & 0.60 & \second{0.43} \\
\midrule


\multirow{3}{*}{\makecell[c]{Hypernet.}}
  & SHINE
  & 88.6 & \second{69.2} & 59.3 & 6.25 & 36.0
  & \second{70.8} & \second{59.3}
  & \second{29.8} & \second{0.53} & 0.50 \\

  & Text-to-LoRA
  & 77.1 & 46.2 & \second{63.0} & 50.0 & 24.0 & 29.2
  & 50.7 & 37.1 & 0.59 & 0.44 \\

  & \textbf{\method{}}
  & \best{97.1} & \best{92.3} & \second{63.0} & 43.8
  & \best{64.0} & \best{75.0} & \best{74.3}
  & \best{28.4} & \best{0.44} & \best{0.34} \\

\specialrule{\lightrulewidth}{\aboverulesep}{0pt}
\rowcolor{black!7}
\multicolumn{12}{c}{\textit{Unseen Split}} \\
\specialrule{\lightrulewidth}{0pt}{\belowrulesep}

\multirow{5}{*}{\makecell[c]{Prompt}}
  & Vanilla
  & 54.2 & 55.6 & 41.9 & \second{47.8} & 57.1 & 23.5
  & 47.0 & 34.9 & \best{0.44} & 0.62 \\

  & Few-Shot
  & 66.7 & 50.0 & 61.3 & \best{56.5} & \second{76.2} & 0.00
  & 54.5 & \best{28.5} & 2.04 & 0.50 \\

  & Reflexion
  & 50.0 & 61.1 & 58.1 & 43.5 & 66.7 & 0.00
  & 48.5 & 32.8 & 0.57 & 0.82 \\

  & AdaPlanner
  & \second{83.3} & 50.0 & 54.8 & 39.1 & 52.4 & 29.4
  & 53.0 & 34.3 & 0.54 & 0.75 \\

  & In-Context Skill
  & 70.8 & 61.1 & \best{74.2} & 43.5 & 47.6 & 23.5
  & 56.0 & \second{29.7} & 1.23 & 0.61 \\
\midrule

\multirow{3}{*}{\makecell[c]{Fine-tuning}}
  & Full SFT
  & 54.2 & 50.0 & 58.1 & 43.5 & 61.9 & 35.3
  & 51.5 & 37.2 & 1.34 & 0.48 \\

  & Shared LoRA
  & 54.2 & 38.9 & 58.1 & \second{47.8} & 42.9 & 35.3
  & 47.8 & 41.9 & 1.33 & 0.50 \\

  & Per-skill LoRA
  & 75.0 & 33.3 & 48.4 & 39.1 & 52.4 & 47.1
  & 50.0 & 42.2 & 0.57 & \best{0.46} \\
\midrule

\multirow{3}{*}{\makecell[c]{Hypernet.}}
  & SHINE
  & 75.0 & \best{72.2} & \second{71.0} & 43.5 & 38.1
  & \second{64.7} & \second{61.2}
  & 32.1 & \second{0.49} & 0.53 \\

  & Text-to-LoRA
  & 66.7 & 50.0 & 45.2 & 34.8 & 52.4 & 35.3
  & 47.8 & 45.5 & 0.56 & \second{0.47} \\

  & \textbf{\method{}}
  & \best{91.7} & \second{66.7} & 64.5 & 43.5
  & \best{81.0} & \best{70.6} & \best{69.4}
  & 31.4 & \best{0.44} & 0.51 \\

\bottomrule
\end{tabularx}

\caption{
Performance on \textbf{ALFWorld} in success rate (\%).
Results are reported on the \emph{seen} and \emph{unseen}
splits with a per-task breakdown, and \emph{Avg} denotes the micro-averaged success rate over all evaluation episodes.
\emph{Step} denotes the average number of interaction steps
per episode, while \emph{Prefill} and \emph{Decode} denote
the average token counts (k) per step.
The best and second-best results are highlighted in
\textbf{bold} and \underline{underlined}, respectively.
}
\label{tab:alfworld}
\end{table*}

\newlength{\searchcategorywidth}
\newlength{\searchmethodwidth}
\setlength{\searchcategorywidth}{1.85cm}
\setlength{\searchmethodwidth}{2.20cm}

\begin{table*}[t]
\centering
\small
\setlength{\tabcolsep}{2.5pt}
\renewcommand{\arraystretch}{1.05}

\begin{tabularx}{\textwidth}{
  @{}
  >{\centering\arraybackslash}p{\searchcategorywidth}
  >{\raggedright\arraybackslash}p{\searchmethodwidth}
  *{9}{>{\centering\arraybackslash}X}
  @{}
}
\toprule
\multirow{2}{*}{\textbf{Category}}
  & \multirow{2}{*}{\textbf{Method}}
  & \multicolumn{3}{c}{\textbf{Single-Hop QA}}
  & \multicolumn{4}{c}{\textbf{Multi-Hop QA}}
  & \multirow{2}{*}{\textbf{Avg}$\uparrow$}
  & \multirow{2}{*}{\makecell{\textbf{Cost}$\downarrow$}} \\
\cmidrule(lr){3-5}
\cmidrule(lr){6-9}
  & 
  & \textbf{NQ}$^\dagger$
  & \textbf{Triv}$^\star$
  & \textbf{Pop}$^\star$
  & \textbf{Hotp}$^\dagger$
  & \textbf{2WK}$^\star$
  & \textbf{MuS}$^\star$
  & \textbf{Bam}$^\star$
  & & \\
\midrule

\multirow{6}{*}{\makecell[c]{Prompt}}
  & Vanilla
  & 25.2 & 50.6 & 35.2 & 26.2 & 26.8 & 4.20 & 28.8
  & 28.1 & \second{0.24} \\

  & CoT
  & 19.2 & 50.8 & 20.0 & 22.4 & 26.2 & 5.00
  & \second{37.6} & 24.5 & \best{0.09} \\

  & Few-Shot
  & 34.6 & 57.6 & 39.4 & 30.0 & 25.8 & 6.00 & 17.6
  & 31.7 & 0.94 \\

  & R1-Instruct
  & 27.0 & 55.0 & 31.6 & 26.6 & \second{33.6} & 5.80 & 34.4
  & 30.1 & \second{0.24} \\

  & RAG
  & \best{39.0} & \best{64.0} & \best{45.0} & 32.4
  & 21.2 & 6.80 & 27.2
  & \second{34.4} & 0.89 \\

  & In-Context Skill
  & 27.2 & 56.4 & 33.0 & 30.2 & \best{39.8} & 7.60
  & \best{38.4} & 32.6 & 1.10 \\
\midrule

\multirow{3}{*}{\makecell[c]{Fine-tuning}}
  & Full SFT
  & 35.4 & \second{58.0} & 39.8 & \second{38.4}
  & 25.8 & 10.2 & 24.8
  & 34.2 & 1.52 \\

  & Shared LoRA
  & 35.4 & 55.0 & 40.6 & 38.0
  & 27.2 & 9.20 & 22.4
  & 33.8 & 1.49 \\

  & Per-skill LoRA
  & 32.8 & 56.6 & 39.0 & 37.6
  & 24.0 & \second{10.4} & 16.8
  & 32.8 & 0.51 \\
\midrule

\multirow{3}{*}{\makecell[c]{Hypernet.}}
  & SHINE
  & 34.6 & 57.4 & 35.6 & 36.2
  & 29.4 & \best{12.4} & 30.4
  & 34.1 & 0.39 \\

  & Text-to-LoRA
  & 32.8 & 43.2 & 25.2 & 30.2
  & 30.4 & 8.00 & 12.8
  & 27.7 & 0.42 \\

  & \textbf{\method{}}
  & \second{36.2} & 57.6 & \second{41.0} & \best{39.6}
  & 32.0 & 9.80 & 25.6
  & \best{35.6} & 0.31 \\

\bottomrule
\end{tabularx}

\caption{
Performance on \textbf{Search-QA} in exact match (\%).
Avg is micro-averaged over all examples. Each dataset contains 500 evaluation examples, except Bamboogle, which contains 125.
$\dagger$ and $\star$ indicate in-domain and out-of-domain
datasets, respectively.
\emph{Cost} denotes the average token count (k) per step.
The best and second-best results are highlighted in
\textbf{bold} and \underline{underlined}, respectively.
}
\label{tab:search}
\end{table*}

\paragraph{Method Performance.}
Tables~\ref{tab:alfworld} and~\ref{tab:search} show that \method{} achieves the best average performance on both benchmarks. On ALFWorld, it reaches average success rates of 74.3\% and 69.4\% on the seen and unseen splits. It outperforms SHINE by 15.0 and 8.2 percentage points and In-Context Skill by 21.4 and 13.4 points on the two splits, respectively. On Search-QA, \method{} achieves the highest average EM of 35.6\%, outperforming RAG by 1.2 points and SHINE by 1.5 points. It ranks first on HotpotQA and second on NQ and PopQA.

\paragraph{Cross-Backbone Generalization.}
\begin{table}[t]
\centering
\small
\setlength{\tabcolsep}{5pt}
\renewcommand{\arraystretch}{1.08}
\begin{tabular}{lccc}
\toprule
& \multicolumn{2}{c}{\textbf{ALFWorld SR} ($\uparrow$)}
& \textbf{Search-QA EM} ($\uparrow$) \\
\cmidrule(lr){2-3}\cmidrule(lr){4-4}
\textbf{Method} & \textbf{Seen} & \textbf{Unseen} & \textbf{Avg.} \\
\midrule
Vanilla (Llama3-8B)          & 22.1 & 18.7 & 16.6 \\
\midrule
Reflexion        & 17.1 & 17.9 & --    \\
AdaPlanner       &  5.7 &  7.5 & --    \\
CoT              & --    & --    & 19.2 \\
R1-Instruct      & --    & --    & 20.0 \\
RAG              & --    & --    & 24.0 \\
\midrule
In-Context Skill & 25.0 & 23.9 & 24.5    \\
\textbf{\method{}} & \textbf{40.7} & \textbf{31.3}
                   & \textbf{25.0} \\
\bottomrule
\end{tabular}
\caption{
Results on Llama3-8B. ALFWorld reports the average success rate (SR)
over six task categories, while Search-QA reports the average exact match
(EM) over seven datasets. ``--'' denotes results that have not been evaluated.
}
\label{tab:llama3_results}
\end{table}
To examine whether the skill-oriented training pipeline generalizes beyond Qwen3-8B, we further apply the same two-stage training procedure to Llama3-8B. As shown in Table~\ref{tab:llama3_results}, \method{}
outperforms In-Context Skill by 15.7 and 7.4 percentage points on the ALFWorld seen and unseen splits, respectively. On Search-QA, it achieves an average EM of 25.0\%. These results suggest that the benefits of the skill-oriented training and deployment pipeline are not limited to the Qwen3-8B backbone.

\paragraph{Out-of-Domain Generalization.}
We further evaluate \method{} on WebShop \citep{yao2023webshopscalablerealworldweb} using textual skills from
\citet{Xia2026SkillRL}, while excluding all WebShop trajectories from
training. As shown in Table~\ref{tab:webshop_ood}, the pretrained and SFT checkpoints achieve scores of 18.5 and 23.9, compared with 10.3 for Vanilla and 6.7 for In-Context Skill. 
The In-Context result suggests that
long skill prompts may dilute attention to the current state and
admissible actions, highlighting the advantage of compiling skills
into latent space.

\begin{table}[t]
\centering
\small
\setlength{\tabcolsep}{7pt}
\renewcommand{\arraystretch}{1.05}
\begin{tabular}{lcc}
\toprule
\textbf{Method}
  & \textbf{Score}$\uparrow$
  & \textbf{Step}$\downarrow$ \\
\midrule
Vanilla
  & 10.3 & 24.8 \\
In-Context Skill
  & 6.7 & 27.4 \\
\method{} (Pretrain)
  & \second{18.5} & \best{21.4} \\
\method{} (SFT)
  & \best{23.9} & \second{21.7} \\
\bottomrule
\end{tabular}
\caption{
WebShop OOD results without training trajectories. \emph{Score} denotes the standard WebShop reward scaled by 100.
}
\label{tab:webshop_ood}
\end{table}

\paragraph{Effect of Teacher Supervision and Hypernetwork.}
Compared with Full SFT and Shared LoRA, \method{} improves average performance by 20.7 and 29.3 percentage points on ALFWorld seen, by 17.9 and 21.6 points on ALFWorld unseen, and by 1.4 and 1.8 points on Search-QA. These supervision-matched comparisons show that teacher trajectories alone do not account for the gains. Compared with Per-skill LoRA, \method{} improves performance by 28.6, 19.4, and 2.8 points on ALFWorld seen, ALFWorld unseen, and Search-QA, respectively. This comparison supports the benefit of shared hypernetwork generation over independent skill-specific optimization. Its consistent advantage over SHINE and Text-to-LoRA further suggests that the improvement does not arise merely from the use of a hypernetwork.

\paragraph{Token Efficiency.}
Compared with In-Context Skill, \method{} reduces prefill cost by 63.6\% and 64.2\% on the ALFWorld seen and unseen splits and reduces the average per-step token cost by 71.8\% on Search-QA. Meanwhile, it improves ALFWorld success rates by 21.4 and 13.4 percentage points and Search-QA EM by 3.0 points. Among the fine-tuning and hypernetwork baselines, \method{} achieves the lowest ALFWorld prefill cost and Search-QA per-step token count.

Beyond cost, \method{} also shortens interaction trajectories. On the seen split, it achieves the fewest average steps per episode, reducing the trajectory length from 35.0 steps for the Vanilla backbone to 28.4 steps. This indicates that latent skill weights can help the agent reach successful outcomes with fewer environment interactions, rather than merely improving final success rates.

Tables~\ref{tab:alfworld} and~\ref{tab:search} show that \method{} improves both performance and token efficiency over explicit text injection. We next ask whether the resulting skill weight space has structure that can be analyzed and used. We find that \method{} exhibits three useful properties: \emph{structured}, \emph{controllable}, and \emph{composable}.


\subsection{Structured: Semantic Geometry of the LoRA Weight Space}
To examine whether \method{} forms a meaningful latent geometry, we apply Multidimensional Scaling (MDS) to the LoRA weights of the 8 in-domain skills at both the pretrain and SFT stages. We also evaluate out-of-distribution (OOD) skill texts collected from public GitHub repositories, covering Code, Finance, and Writing. 

\begin{figure}[t]
  \centering
  \includegraphics[width=\columnwidth]{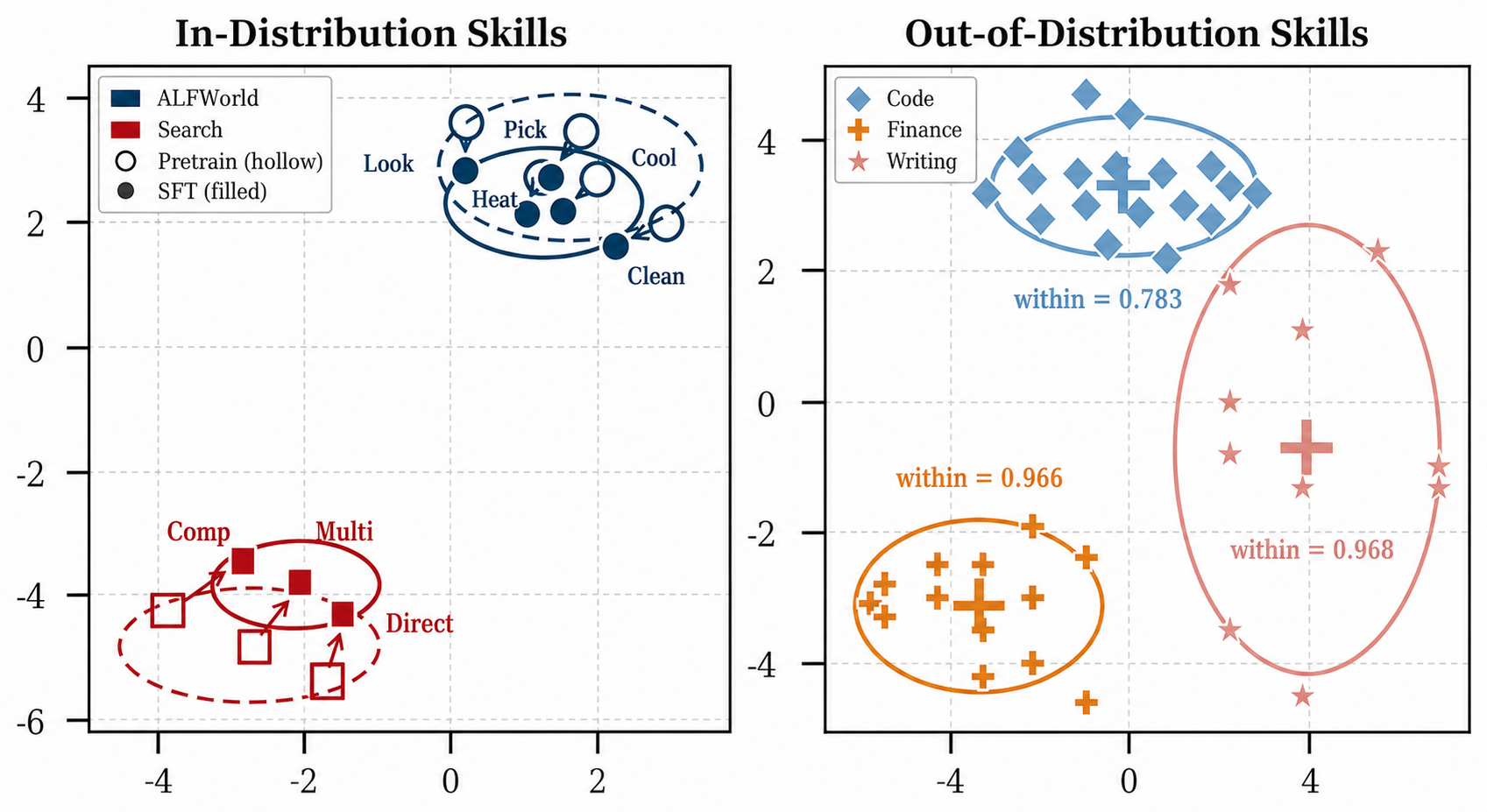}
  \caption{
    MDS visualization of LoRA weights.
    ``$+$'' marks each cluster centroid; \emph{within} reports mean intra-cluster cosine similarity. Axes are scaled by $\times 10^{-2}$.
  }
  \label{fig:lora_mds}
\end{figure}

As shown in Figure~\ref{fig:lora_mds} (left), the 5 ALFWorld and 3 Search skills form distinct domain clusters, with an inter-cluster distance of 0.0887 and higher within-domain than cross-domain similarity (0.982 vs.\ 0.910). After SFT, the distance decreases by 20.6\% to 0.0704 while both similarities increase, indicating shared agent behaviors alongside preserved skill structure. Figure~\ref{fig:lora_mds} (right) shows similar organization for OOD skills: Code, Finance, and Writing form distinct clusters with within-domain similarities of 0.783, 0.9664, and 0.9681, respectively, all exceeding their cross-domain similarities. Despite the lack of domain-specific supervision, these patterns suggest that the compiler maps procedural text into a semantically structured weight space rather than overfitting to in-domain skills. Given this geometric structure, a natural question arises: can we leverage this structure to actively control the injection strength of a skill?
 
\subsection{Controllable: Continuous Modulation of Skill Injection Strength}

We evaluate continuous control of skill injection strength by scaling the hypernetwork output $\Delta W$ with a coefficient $\alpha$. We test nine values from 0 to 1.2 on the ALFWorld seen and unseen splits. Here, $\alpha{=}0$ removes the generated LoRA, while $\alpha{=}1$ uses it without rescaling.

\begin{figure}[t]
  \centering
  \includegraphics[width=\columnwidth]
  {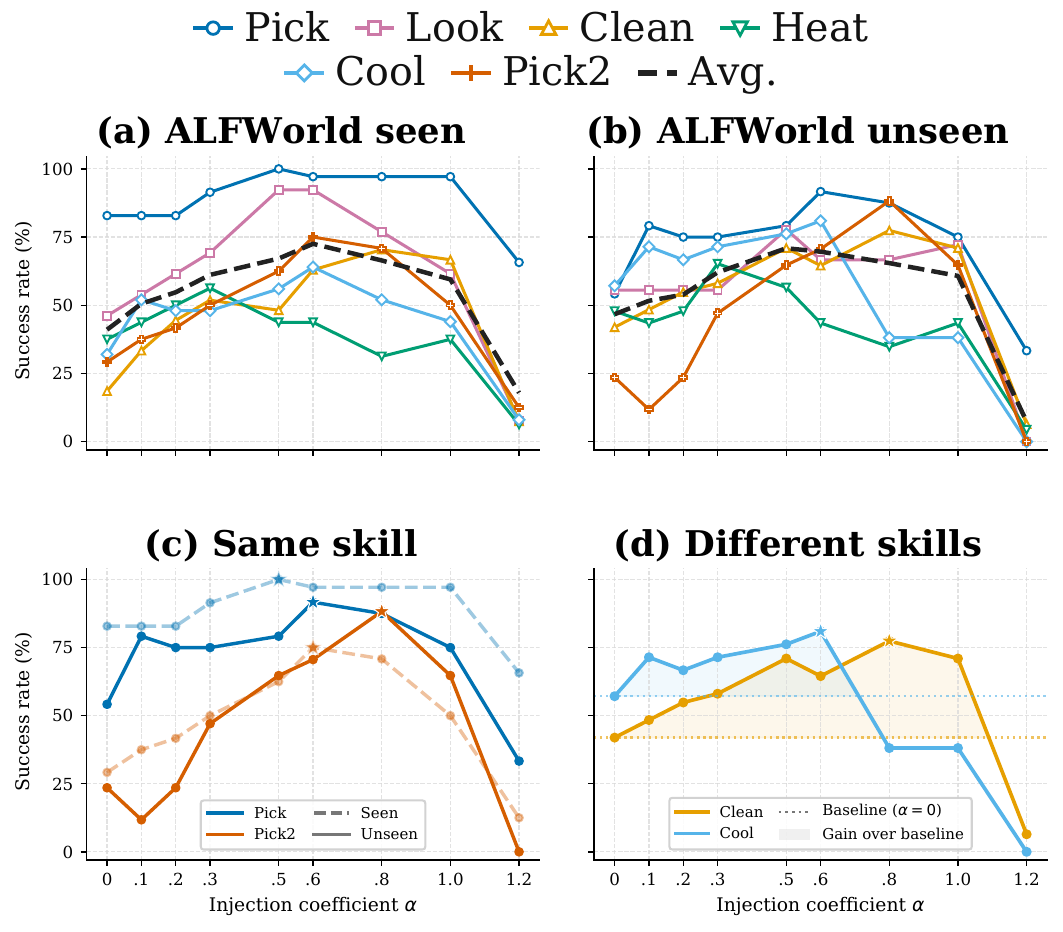}
  \caption{
    ALFWorld performance under different injection coefficients
    $\alpha$.
    \textbf{(a--b)} Per-task and average results on seen and unseen
    splits.
    \textbf{(c)} Pick and Pick2 share the same skill.
    \textbf{(d)} Clean and Cool use different skills on the unseen
    split.
  }
  \label{fig:scale}
\end{figure}

Figure~\ref{fig:scale}(a--b) shows an inverted-U trend in average
performance. Success rises from 43.57\% to 74.29\% at $\alpha{=}0.6$
on the seen split and from 47.01\% to 70.90\% at $\alpha{=}0.5$ on
the unseen split, but falls to 22.86\% and 8.21\% at $\alpha{=}1.2$.
The task-specific optima are identical or adjacent across splits,
indicating a stable effective injection range.
The optimal coefficient nevertheless varies across tasks. Applying
the seen-optimal coefficient $\alpha{=}0.6$ to the unseen split falls
short of the task-specific optima by 21.74, 17.65, and 12.90 points
on Heat, Pick2, and Clean, respectively. Figure~\ref{fig:scale}(c--d)
further shows that Pick2 and Clean, which have weaker backbone
performance than Pick and Cool, prefer stronger injection. These results demonstrate continuous skill control
and motivate adaptive selection of $\alpha$.

\subsection{Composable: Skill Arithmetic in Parameter Space}

We next examine whether \method{} supports parameter-space skill composition. We compose \emph{Look} with the complementary \emph{Pick} skill and compare single-skill adapters with Direct, Text, and Component Merging on all 31 Look episodes. Successful composition should introduce Pick behavior without impairing Look capability.

As shown in Table~\ref{tab:compose}, Component Merging performs best on both splits, achieving 84.6\% seen and 77.8\% unseen success rates. It adds three seen and one unseen success over Look-Only without losing any original success. By including shared general and mistake-avoidance components only once, it avoids the duplicated weights of Direct Merging and the out-of-distribution input of Text Merging. These results show that reliable LoRA composition requires semantically aligned components. Case studies are provided in Appendix.



\begin{table}[t]
\centering
\small
\setlength{\tabcolsep}{5pt}
\renewcommand{\arraystretch}{1.1}
\begin{tabular}{l cc cc}
\toprule
& \multicolumn{2}{c}{\textbf{Seen}}
& \multicolumn{2}{c}{\textbf{Unseen}} \\
\cmidrule(lr){2-3}\cmidrule(lr){4-5}
\textbf{Method}
  & \textbf{Ep.} & \textbf{\%}
  & \textbf{Ep.} & \textbf{\%} \\
\midrule
Look-Only      & 8/13  & 61.5        & 13/18 & 72.2 \\
Pick-Only      & 8/13  & 61.5        & 11/18 & 61.1 \\
\midrule
Direct Merging & 9/13  & 69.2        & 11/18 & 61.1 \\
Text Merging   & 8/13  & 61.5        & 11/18 & 61.1 \\
\textbf{Component Merging} & 11/13 & \best{84.6} & 14/18 & \best{77.8} \\
\bottomrule
\end{tabular}
\caption{
  Skill composition results on all 31 Look task episodes under five skill
composition configurations.
}
\label{tab:compose}

\end{table}

\subsection{Robustness and Security}
We evaluate four skill-text perturbations, including paraphrasing, Markdown removal, section reordering, and noise injection. We also consider two prompt-level attacks targeting instruction hijacking and skill extraction. As shown in Table~\ref{tab:sensitivity_security}, \method{} maintains a consistent advantage over In-context Skill under all perturbations. On ALFWorld, the margin ranges from 17.2 to 24.3 points, comparable to the 21.4-point margin under the base condition. On Search-QA, the latent-skill approach also remains ahead under every perturbation. Notably, removing Markdown formatting causes no degradation on ALFWorld, suggesting that the generated LoRA captures skill semantics rather than relying on surface formatting. Across the four perturbations, the average success rate is 70.7\%, only 3.6 points below the base performance.

The same pattern holds under prompt-level attacks. Under Hijack, In-context Skill drops from 52.9\% to 8.57\% on ALFWorld, while the weight-space variant retains 38.6\%; on Search-QA, \method{} drops by only 1.6 points. Under Extract, In-context Skill is vulnerable because the skill text is directly present in the prompt, whereas weight-space storage reduces direct plaintext exposure. These results suggest that moving skills from prompt space to LoRA weights improves not only efficiency, but also robustness to prompt perturbations and prompt-level attacks.

\begin{table}[t]
\centering
\small
\setlength{\tabcolsep}{5pt}
\renewcommand{\arraystretch}{1.1}
\begin{tabular}{l c>{}c c>{}c}
\toprule
& \multicolumn{2}{c}{\textbf{ALFWorld}}
& \multicolumn{2}{c}{\textbf{Search-QA}} \\
\cmidrule(lr){2-3}\cmidrule(lr){4-5}
\textbf{Method}
  & \textbf{In-context} & \textbf{Latent}
  & \textbf{In-context} & \textbf{Latent} \\
\midrule
Base       & 52.9 & \best{74.3} & 32.6 & \best{35.6} \\
\midrule
Paraphrase & 50.7 & \best{67.9} & 33.2 & \best{34.0} \\
Plaintext  & 50.0 & \best{74.3} & 32.4 & \best{34.4} \\
Reorder    & 50.7 & \best{69.3} & 32.4 & \best{33.6} \\
Noise      & 47.9 & \best{71.4} & 31.7 & \best{33.7} \\
\midrule
Hijack     & 8.57 & \best{38.6} & 23.5 & \best{34.0} \\
Extract    & 48.6 & \best{70.0} & 21.3 & \best{29.3} \\
\bottomrule
\end{tabular}
\caption{
  Average performance under skill text perturbations and adversarial attacks.
  ALFWorld reports success rate (\%) and Search-QA reports exact match (\%).
}
\label{tab:sensitivity_security}

\end{table}

\section{Conclusion}
\method{} adopts the text-to-LoRA hypernetwork architecture and trains it for agent skill encoding through large-scale skill-document pretraining and trajectory-supervised fine-tuning. The resulting LoRAs move reusable procedural knowledge from context space into weight space and
can be mounted without including the original skill text in the prompt.
Across ALFWorld and Search-QA, this design improves over direct in-context skill prompting while substantially reducing the repeated prefill overhead introduced by skill text. Beyond efficiency, our analyses show that the generated skill LoRAs form a structured semantic geometry, can be controlled through the injection coefficient, and can be composed in parameter space when skill components are properly aligned. These results suggest that latent skill weights offer a practical substrate for building LLM agents whose skills are efficient, modular, controllable, and less directly exposed as plaintext prompts.



\bibliography{aaai2027}

@misc{Yao2023ReAct,
      title={ReAct: Synergizing Reasoning and Acting in Language Models}, 
      author={Shunyu Yao and Jeffrey Zhao and Dian Yu and Nan Du and Izhak Shafran and Karthik Narasimhan and Yuan Cao},
      year={2023},
      eprint={2210.03629},
      archivePrefix={arXiv},
      primaryClass={cs.CL},
      url={https://arxiv.org/abs/2210.03629}, 
}

@inproceedings{Shinn2023Reflexion,
    author = {Shinn, Noah and Cassano, Federico and Gopinath, Ashwin and Narasimhan, Karthik and Yao, Shunyu},
    title = {Reflexion: language agents with verbal reinforcement learning},
    year = {2023},
    publisher = {Curran Associates Inc.},
    address = {Red Hook, NY, USA},
    booktitle = {Proceedings of the 37th International Conference on Neural Information Processing Systems},
    articleno = {377},
    pages = {8634--8652},
    location = {New Orleans, LA, USA},
    series = {NIPS '23}
}

@misc{Wang2023Voyager,
      title={Voyager: An Open-Ended Embodied Agent with Large Language Models}, 
      author={Guanzhi Wang and Yuqi Xie and Yunfan Jiang and Ajay Mandlekar and Chaowei Xiao and Yuke Zhu and Linxi Fan and Anima Anandkumar},
      year={2023},
      eprint={2305.16291},
      archivePrefix={arXiv},
      primaryClass={cs.AI},
      url={https://arxiv.org/abs/2305.16291}, 
}

@misc{Xia2026SkillRL,
      title={SkillRL: Evolving Agents via Recursive Skill-Augmented Reinforcement Learning}, 
      author={Peng Xia and Jianwen Chen and Hanyang Wang and Jiaqi Liu and Kaide Zeng and Yu Wang and Siwei Han and Yiyang Zhou and Xujiang Zhao and Haifeng Chen and Zeyu Zheng and Cihang Xie and Huaxiu Yao},
      year={2026},
      eprint={2602.08234},
      archivePrefix={arXiv},
      primaryClass={cs.LG},
      url={https://arxiv.org/abs/2602.08234}, 
}

@misc{Wu2025EvolveR,
      title={EvolveR: Self-Evolving LLM Agents through an Experience-Driven Lifecycle}, 
      author={Rong Wu and Xiaoman Wang and Jianbiao Mei and Pinlong Cai and Daocheng Fu and Cheng Yang and Licheng Wen and Xuemeng Yang and Yufan Shen and Yuxin Wang and Botian Shi},
      year={2026},
      eprint={2510.16079},
      archivePrefix={arXiv},
      primaryClass={cs.CL},
      url={https://arxiv.org/abs/2510.16079}, 
}

@misc{SkillOS2026,
      title={SkillOS: Learning Skill Curation for Self-Evolving Agents}, 
      author={Siru Ouyang and Jun Yan and Yanfei Chen and Rujun Han and Zifeng Wang and Bhavana Dalvi Mishra and Rui Meng and Chun-Liang Li and Yizhu Jiao and Kaiwen Zha and Maohao Shen and Vishy Tirumalashetty and George Lee and Jiawei Han and Tomas Pfister and Chen-Yu Lee},
      year={2026},
      eprint={2605.06614},
      archivePrefix={arXiv},
      primaryClass={cs.AI},
      url={https://arxiv.org/abs/2605.06614}, 
}

@inproceedings{Zhao2024ExpeL,
author = {Zhao, Andrew and Huang, Daniel and Xu, Quentin and Lin, Matthieu and Liu, Yong-Jin and Huang, Gao},
title = {ExpeL: LLM agents are experiential learners},
year = {2024},
isbn = {978-1-57735-887-9},
publisher = {AAAI Press},
url = {https://doi.org/10.1609/aaai.v38i17.29936},
doi = {10.1609/aaai.v38i17.29936},
booktitle = {Proceedings of the Thirty-Eighth AAAI Conference on Artificial Intelligence and Thirty-Sixth Conference on Innovative Applications of Artificial Intelligence and Fourteenth Symposium on Educational Advances in Artificial Intelligence},
articleno = {2188},
pages = {19632--19642},
series = {AAAI'24/IAAI'24/EAAI'24}
}

@misc{Chen2023FireAct,
      title={FireAct: Toward Language Agent Fine-tuning}, 
      author={Baian Chen and Chang Shu and Ehsan Shareghi and Nigel Collier and Karthik Narasimhan and Shunyu Yao},
      year={2023},
      eprint={2310.05915},
      archivePrefix={arXiv},
      primaryClass={cs.CL},
      url={https://arxiv.org/abs/2310.05915}, 
}

@inproceedings{Zeng2024AgentTuning,
    title = "{A}gent{T}uning: Enabling Generalized Agent Abilities for {LLM}s",
    author = "Zeng, Aohan  and
      Liu, Mingdao  and
      Lu, Rui  and
      Wang, Bowen  and
      Liu, Xiao  and
      Dong, Yuxiao  and
      Tang, Jie",
    editor = "Ku, Lun-Wei  and
      Martins, Andre  and
      Srikumar, Vivek",
    booktitle = "Findings of the Association for Computational Linguistics: ACL 2024",
    month = aug,
    year = "2024",
    address = "Bangkok, Thailand",
    publisher = "Association for Computational Linguistics",
    url = "https://aclanthology.org/2024.findings-acl.181/",
    doi = "10.18653/v1/2024.findings-acl.181",
    pages = "3053--3077"
}

@misc{Lu2026Skill0,
      title={SKILL0: In-Context Agentic Reinforcement Learning for Skill Internalization}, 
      author={Zhengxi Lu and Zhiyuan Yao and Jinyang Wu and Chengcheng Han and Qi Gu and Xunliang Cai and Weiming Lu and Jun Xiao and Yueting Zhuang and Yongliang Shen},
      year={2026},
      eprint={2604.02268},
      archivePrefix={arXiv},
      primaryClass={cs.LG},
      url={https://arxiv.org/abs/2604.02268}, 
}

@inproceedings{Greshake2023IndirectInjection,
author = {Greshake, Kai and Abdelnabi, Sahar and Mishra, Shailesh and Endres, Christoph and Holz, Thorsten and Fritz, Mario},
title = {Not What You've Signed Up For: Compromising Real-World LLM-Integrated Applications with Indirect Prompt Injection},
year = {2023},
isbn = {9798400702600},
publisher = {Association for Computing Machinery},
address = {New York, NY, USA},
url = {https://doi.org/10.1145/3605764.3623985},
doi = {10.1145/3605764.3623985},
booktitle = {Proceedings of the 16th ACM Workshop on Artificial Intelligence and Security},
pages = {79--90},
numpages = {12},
location = {Copenhagen, Denmark},
series = {AISec '23}
}

@misc{Liu2023PromptInjection,
      title={Prompt Injection attack against LLM-integrated Applications}, 
      author={Yi Liu and Gelei Deng and Yuekang Li and Kailong Wang and Zihao Wang and Xiaofeng Wang and Tianwei Zhang and Yepang Liu and Haoyu Wang and Yan Zheng and Leo Yu Zhang and Yang Liu},
      year={2025},
      eprint={2306.05499},
      archivePrefix={arXiv},
      primaryClass={cs.CR},
      url={https://arxiv.org/abs/2306.05499}, 
}

@misc{Schmotz2026SkillSecurity,
      title={Towards Secure Agent Skills: Architecture, Threat Taxonomy, and Security Analysis}, 
      author={Zhiyuan Li and Jingzheng Wu and Xiang Ling and Xing Cui and Tianyue Luo},
      year={2026},
      eprint={2604.02837},
      archivePrefix={arXiv},
      primaryClass={cs.CR},
      url={https://arxiv.org/abs/2604.02837}, 
}

@misc{Liu2026SHINE,
      title={SHINE: A Scalable In-Context Hypernetwork for Mapping Context to LoRA in a Single Pass}, 
      author={Yewei Liu and Xiyuan Wang and Yansheng Mao and Yoav Gelbery and Haggai Maron and Muhan Zhang},
      year={2026},
      eprint={2602.06358},
      archivePrefix={arXiv},
      primaryClass={cs.CL},
      url={https://arxiv.org/abs/2602.06358}, 
}

@misc{Ha2017Hypernetwork,
      title={HyperNetworks}, 
      author={David Ha and Andrew Dai and Quoc V. Le},
      year={2016},
      eprint={1609.09106},
      archivePrefix={arXiv},
      primaryClass={cs.LG},
      url={https://arxiv.org/abs/1609.09106}, 
}

@misc{Hu2022LoRA,
      title={LoRA: Low-Rank Adaptation of Large Language Models}, 
      author={Edward J. Hu and Yelong Shen and Phillip Wallis and Zeyuan Allen-Zhu and Yuanzhi Li and Shean Wang and Lu Wang and Weizhu Chen},
      year={2021},
      eprint={2106.09685},
      archivePrefix={arXiv},
      primaryClass={cs.CL},
      url={https://arxiv.org/abs/2106.09685}, 
}

@misc{Chhikara2025Mem0,
      title={Mem0: Building Production-Ready AI Agents with Scalable Long-Term Memory}, 
      author={Prateek Chhikara and Dev Khant and Saket Aryan and Taranjeet Singh and Deshraj Yadav},
      year={2025},
      eprint={2504.19413},
      archivePrefix={arXiv},
      primaryClass={cs.CL},
      url={https://arxiv.org/abs/2504.19413}, 
}

@misc{Chen2025GenAdapter,
      title={Generative Adapter: Contextualizing Language Models in Parameters with A Single Forward Pass}, 
      author={Tong Chen and Hao Fang and Patrick Xia and Xiaodong Liu and Benjamin Van Durme and Luke Zettlemoyer and Jianfeng Gao and Hao Cheng},
      year={2024},
      eprint={2411.05877},
      archivePrefix={arXiv},
      primaryClass={cs.LG},
      url={https://arxiv.org/abs/2411.05877}, 
}

@misc{Zhou2026Externalization,
      title={Externalization in LLM Agents: A Unified Review of Memory, Skills, Protocols and Harness Engineering}, 
      author={Chenyu Zhou and Huacan Chai and Wenteng Chen and Zihan Guo and Rong Shan and Yuanyi Song and Tianyi Xu and Yingxuan Yang and Aofan Yu and Weiming Zhang and Congming Zheng and Jiachen Zhu and Zeyu Zheng and Zhuosheng Zhang and Xingyu Lou and Changwang Zhang and Zhihui Fu and Jun Wang and Weiwen Liu and Jianghao Lin and Weinan Zhang},
      year={2026},
      eprint={2604.08224},
      archivePrefix={arXiv},
      primaryClass={cs.SE},
      url={https://arxiv.org/abs/2604.08224}, 
}

@misc{Cho2026SkillRet,
      title={SkillRet: A Large-Scale Benchmark for Skill Retrieval in LLM Agents}, 
      author={Hongcheol Cho and Ryangkyung Kang and Youngeun Kim},
      year={2026},
      eprint={2605.05726},
      archivePrefix={arXiv},
      primaryClass={cs.AI},
      url={https://arxiv.org/abs/2605.05726}, 
}

@inproceedings{Jiang2023LLMLingua,
    title = "{LLML}ingua: Compressing Prompts for Accelerated Inference of Large Language Models",
    author = "Jiang, Huiqiang  and
      Wu, Qianhui  and
      Lin, Chin-Yew  and
      Yang, Yuqing  and
      Qiu, Lili",
    editor = "Bouamor, Houda  and
      Pino, Juan  and
      Bali, Kalika",
    booktitle = "Proceedings of the 2023 Conference on Empirical Methods in Natural Language Processing",
    month = dec,
    year = "2023",
    address = "Singapore",
    publisher = "Association for Computational Linguistics",
    url = "https://aclanthology.org/2023.emnlp-main.825/",
    doi = "10.18653/v1/2023.emnlp-main.825",
    pages = "13358--13376",
}

@inproceedings{Jiang2024LongLLMLingua,
    title = "{L}ong{LLML}ingua: Accelerating and Enhancing {LLM}s in Long Context Scenarios via Prompt Compression",
    author = "Jiang, Huiqiang  and
      Wu, Qianhui  and
      Luo, Xufang  and
      Li, Dongsheng  and
      Lin, Chin-Yew  and
      Yang, Yuqing  and
      Qiu, Lili",
    editor = "Ku, Lun-Wei  and
      Martins, Andre  and
      Srikumar, Vivek",
    booktitle = "Proceedings of the 62nd Annual Meeting of the Association for Computational Linguistics (Volume 1: Long Papers)",
    month = aug,
    year = "2024",
    address = "Bangkok, Thailand",
    publisher = "Association for Computational Linguistics",
    url = "https://aclanthology.org/2024.acl-long.91/",
    doi = "10.18653/v1/2024.acl-long.91",
    pages = "1658--1677",
}

@article{Liu2024LostMiddle,
    title = "Lost in the Middle: How Language Models Use Long Contexts",
    author = "Liu, Nelson F.  and
      Lin, Kevin  and
      Hewitt, John  and
      Paranjape, Ashwin  and
      Bevilacqua, Michele  and
      Petroni, Fabio  and
      Liang, Percy",
    journal = "Transactions of the Association for Computational Linguistics",
    volume = "12",
    year = "2024",
    address = "Cambridge, MA",
    publisher = "MIT Press",
    url = "https://aclanthology.org/2024.tacl-1.9/",
    doi = "10.1162/tacl_a_00638",
    pages = "157--173",
}

@misc{Wallace2024InstructionHierarchy,
      title={The Instruction Hierarchy: Training LLMs to Prioritize Privileged Instructions}, 
      author={Eric Wallace and Kai Xiao and Reimar Leike and Lilian Weng and Johannes Heidecke and Alex Beutel},
      year={2024},
      eprint={2404.13208},
      archivePrefix={arXiv},
      primaryClass={cs.CR},
      url={https://arxiv.org/abs/2404.13208}, 
}

@inproceedings{Pfeiffer2021AdapterFusion,
    title = "{A}dapter{F}usion: Non-Destructive Task Composition for Transfer Learning",
    author = {Pfeiffer, Jonas  and
      Kamath, Aishwarya  and
      R{\"u}ckl{\'e}, Andreas  and
      Cho, Kyunghyun  and
      Gurevych, Iryna},
    editor = "Merlo, Paola  and
      Tiedemann, Jorg  and
      Tsarfaty, Reut",
    booktitle = "Proceedings of the 16th Conference of the European Chapter of the Association for Computational Linguistics: Main Volume",
    month = apr,
    year = "2021",
    address = "Online",
    publisher = "Association for Computational Linguistics",
    url = "https://aclanthology.org/2021.eacl-main.39/",
    doi = "10.18653/v1/2021.eacl-main.39",
    pages = "487--503",
}

@misc{Huang2024LoraHub,
      title={LoraHub: Efficient Cross-Task Generalization via Dynamic LoRA Composition}, 
      author={Chengsong Huang and Qian Liu and Bill Yuchen Lin and Tianyu Pang and Chao Du and Min Lin},
      year={2024},
      eprint={2307.13269},
      archivePrefix={arXiv},
      primaryClass={cs.CL},
      url={https://arxiv.org/abs/2307.13269}, 
}

@misc{Shridhar2021ALFWorld,
      title={ALFWorld: Aligning Text and Embodied Environments for Interactive Learning}, 
      author={Mohit Shridhar and Xingdi Yuan and Marc-Alexandre C{\^o}t{\'e} and Yonatan Bisk and Adam Trischler and Matthew Hausknecht},
      year={2021},
      eprint={2010.03768},
      archivePrefix={arXiv},
      primaryClass={cs.CL},
      url={https://arxiv.org/abs/2010.03768}, 
}

@misc{Sun2023AdaPlanner,
      title={AdaPlanner: Adaptive Planning from Feedback with Language Models}, 
      author={Haotian Sun and Yuchen Zhuang and Lingkai Kong and Bo Dai and Chao Zhang},
      year={2023},
      eprint={2305.16653},
      archivePrefix={arXiv},
      primaryClass={cs.CL},
      url={https://arxiv.org/abs/2305.16653}, 
}

@misc{Jin2025SearchR1,
      title={Search-R1: Training LLMs to Reason and Leverage Search Engines with Reinforcement Learning}, 
      author={Bowen Jin and Hansi Zeng and Zhenrui Yue and Jinsung Yoon and Sercan Arik and Dong Wang and Hamed Zamani and Jiawei Han},
      year={2025},
      eprint={2503.09516},
      archivePrefix={arXiv},
      primaryClass={cs.CL},
      url={https://arxiv.org/abs/2503.09516}, 
}

@misc{Wang2022E5,
      title={Text Embeddings by Weakly-Supervised Contrastive Pre-training}, 
      author={Liang Wang and Nan Yang and Xiaolong Huang and Binxing Jiao and Linjun Yang and Daxin Jiang and Rangan Majumder and Furu Wei},
      year={2024},
      eprint={2212.03533},
      archivePrefix={arXiv},
      primaryClass={cs.CL},
      url={https://arxiv.org/abs/2212.03533}, 
}

@misc{Wei2022CoT,
      title={Chain-of-Thought Prompting Elicits Reasoning in Large Language Models}, 
      author={Jason Wei and Xuezhi Wang and Dale Schuurmans and Maarten Bosma and Brian Ichter and Fei Xia and Ed Chi and Quoc Le and Denny Zhou},
      year={2023},
      eprint={2201.11903},
      archivePrefix={arXiv},
      primaryClass={cs.CL},
      url={https://arxiv.org/abs/2201.11903}, 
}

@inproceedings{Lewis2020RAG,
author = {Lewis, Patrick and Perez, Ethan and Piktus, Aleksandra and Petroni, Fabio and Karpukhin, Vladimir and Goyal, Naman and K\"{u}ttler, Heinrich and Lewis, Mike and Yih, Wen-tau and Rockt\"{a}schel, Tim and Riedel, Sebastian and Kiela, Douwe},
title = {Retrieval-augmented generation for knowledge-intensive NLP tasks},
year = {2020},
isbn = {9781713829546},
publisher = {Curran Associates Inc.},
address = {Red Hook, NY, USA},
booktitle = {Proceedings of the 34th International Conference on Neural Information Processing Systems},
articleno = {793},
page = {9459--9474},
location = {Vancouver, BC, Canada},
series = {NIPS '20}
}

@article{Kwiatkowski2019NQ,
    title = "Natural Questions: A Benchmark for Question Answering Research",
    author = "Kwiatkowski, Tom  and
      Palomaki, Jennimaria  and
      Redfield, Olivia  and
      Collins, Michael  and
      Parikh, Ankur  and
      Alberti, Chris  and
      Epstein, Danielle  and
      Polosukhin, Illia  and
      Devlin, Jacob  and
      Lee, Kenton  and
      Toutanova, Kristina  and
      Jones, Llion  and
      Kelcey, Matthew  and
      Chang, Ming-Wei  and
      Dai, Andrew M.  and
      Uszkoreit, Jakob  and
      Le, Quoc  and
      Petrov, Slav",
    editor = "Lee, Lillian  and
      Johnson, Mark  and
      Roark, Brian  and
      Nenkova, Ani",
    journal = "Transactions of the Association for Computational Linguistics",
    volume = "7",
    year = "2019",
    address = "Cambridge, MA",
    publisher = "MIT Press",
    url = "https://aclanthology.org/Q19-1026/",
    doi = "10.1162/tacl_a_00276",
    pages = "452--466"
}

@inproceedings{Joshi2017TriviaQA,
    title = "{T}rivia{QA}: A Large Scale Distantly Supervised Challenge Dataset for Reading Comprehension",
    author = "Joshi, Mandar  and
      Choi, Eunsol  and
      Weld, Daniel  and
      Zettlemoyer, Luke",
    editor = "Barzilay, Regina  and
      Kan, Min-Yen",
    booktitle = "Proceedings of the 55th Annual Meeting of the Association for Computational Linguistics (Volume 1: Long Papers)",
    month = jul,
    year = "2017",
    address = "Vancouver, Canada",
    publisher = "Association for Computational Linguistics",
    url = "https://aclanthology.org/P17-1147/",
    doi = "10.18653/v1/P17-1147",
    pages = "1601--1611"
}

@inproceedings{Mallen2023PopQA,
    title = "When Not to Trust Language Models: Investigating Effectiveness of Parametric and Non-Parametric Memories",
    author = "Mallen, Alex  and
      Asai, Akari  and
      Zhong, Victor  and
      Das, Rajarshi  and
      Khashabi, Daniel  and
      Hajishirzi, Hannaneh",
    editor = "Rogers, Anna  and
      Boyd-Graber, Jordan  and
      Okazaki, Naoaki",
    booktitle = "Proceedings of the 61st Annual Meeting of the Association for Computational Linguistics (Volume 1: Long Papers)",
    month = jul,
    year = "2023",
    address = "Toronto, Canada",
    publisher = "Association for Computational Linguistics",
    url = "https://aclanthology.org/2023.acl-long.546/",
    doi = "10.18653/v1/2023.acl-long.546",
    pages = "9802--9822"
}

@inproceedings{Yang2018HotpotQA,
    title = "{H}otpot{QA}: A Dataset for Diverse, Explainable Multi-hop Question Answering",
    author = "Yang, Zhilin  and
      Qi, Peng  and
      Zhang, Saizheng  and
      Bengio, Yoshua  and
      Cohen, William  and
      Salakhutdinov, Ruslan  and
      Manning, Christopher D.",
    editor = "Riloff, Ellen  and
      Chiang, David  and
      Hockenmaier, Julia  and
      Tsujii, Jun{'}ichi",
    booktitle = "Proceedings of the 2018 Conference on Empirical Methods in Natural Language Processing",
    month = oct # "-" # nov,
    year = "2018",
    address = "Brussels, Belgium",
    publisher = "Association for Computational Linguistics",
    url = "https://aclanthology.org/D18-1259/",
    doi = "10.18653/v1/D18-1259",
    pages = "2369--2380"
}

@inproceedings{Ho2020Wiki,
    title = "Constructing A Multi-hop {QA} Dataset for Comprehensive Evaluation of Reasoning Steps",
    author = "Ho, Xanh  and
      Duong Nguyen, Anh-Khoa  and
      Sugawara, Saku  and
      Aizawa, Akiko",
    editor = "Scott, Donia  and
      Bel, Nuria  and
      Zong, Chengqing",
    booktitle = "Proceedings of the 28th International Conference on Computational Linguistics",
    month = dec,
    year = "2020",
    address = "Barcelona, Spain (Online)",
    publisher = "International Committee on Computational Linguistics",
    url = "https://aclanthology.org/2020.coling-main.580/",
    doi = "10.18653/v1/2020.coling-main.580",
    pages = "6609--6625"
}

@article{Trivedi2022MuSiQue,
    title = "{M}u{S}i{Q}ue: Multihop Questions via Single-hop Question Composition",
    author = "Trivedi, Harsh  and
      Balasubramanian, Niranjan  and
      Khot, Tushar  and
      Sabharwal, Ashish",
    editor = "Roark, Brian  and
      Nenkova, Ani",
    journal = "Transactions of the Association for Computational Linguistics",
    volume = "10",
    year = "2022",
    address = "Cambridge, MA",
    publisher = "MIT Press",
    url = "https://aclanthology.org/2022.tacl-1.31/",
    doi = "10.1162/tacl_a_00475",
    pages = "539--554"
}

@inproceedings{Press2023Bamboogle,
    title = "Measuring and Narrowing the Compositionality Gap in Language Models",
    author = "Press, Ofir  and
      Zhang, Muru  and
      Min, Sewon  and
      Schmidt, Ludwig  and
      Smith, Noah  and
      Lewis, Mike",
    editor = "Bouamor, Houda  and
      Pino, Juan  and
      Bali, Kalika",
    booktitle = "Findings of the Association for Computational Linguistics: EMNLP 2023",
    month = dec,
    year = "2023",
    address = "Singapore",
    publisher = "Association for Computational Linguistics",
    url = "https://aclanthology.org/2023.findings-emnlp.378/",
    doi = "10.18653/v1/2023.findings-emnlp.378",
    pages = "5687--5711"
}

@misc{charakorn2025texttolorainstanttransformeradaption,
      title={Text-to-LoRA: Instant Transformer Adaption}, 
      author={Rujikorn Charakorn and Edoardo Cetin and Yujin Tang and Robert Tjarko Lange},
      year={2025},
      eprint={2506.06105},
      archivePrefix={arXiv},
      primaryClass={cs.LG},
      url={https://arxiv.org/abs/2506.06105}, 
}

@misc{ge2024incontextautoencodercontextcompression,
      title={In-context Autoencoder for Context Compression in a Large Language Model}, 
      author={Tao Ge and Jing Hu and Lei Wang and Xun Wang and Si-Qing Chen and Furu Wei},
      year={2024},
      eprint={2307.06945},
      archivePrefix={arXiv},
      primaryClass={cs.CL},
      url={https://arxiv.org/abs/2307.06945}, 
}

@misc{charakorn2026doctoloralearninginstantlyinternalize,
      title={Doc-to-LoRA: Learning to Instantly Internalize Contexts}, 
      author={Rujikorn Charakorn and Edoardo Cetin and Shinnosuke Uesaka and Robert Tjarko Lange},
      year={2026},
      eprint={2602.15902},
      archivePrefix={arXiv},
      primaryClass={cs.CL},
      url={https://arxiv.org/abs/2602.15902}, 
}

@misc{guo2026skill,
      title={SkillProbe: Security Auditing for Emerging Agent Skill Marketplaces via Multi-Agent Collaboration}, 
      author={Zihan Guo and Zhiyu Chen and Xiaohang Nie and Jianghao Lin and Yuanjian Zhou and Weinan Zhang},
      year={2026},
      eprint={2603.21019},
      archivePrefix={arXiv},
      primaryClass={cs.CR},
      url={https://arxiv.org/abs/2603.21019}, 
}

@misc{pan2026skill,
      title={SkillMAS: Skill Co-Evolution with LLM-based Multi-Agent System}, 
      author={Shuai Pan and Yixiang Liu and Jiaye Gao and Te Gao and Weiwen Liu and Jianghao Lin and Zhihui Fu and Jun Wang and Weinan Zhang and Yong Yu},
      year={2026},
      eprint={2605.09341},
      archivePrefix={arXiv},
      primaryClass={cs.MA},
      url={https://arxiv.org/abs/2605.09341}, 
}

@misc{wang2026skill,
      title={Skills on the Fly: Test-Time Adaptive Skill Synthesis for LLM Agents}, 
      author={Jingxing Wang and Chenyu Zhou and Zhihui Fu and Jun Wang and Weiwen Liu and Weinan Zhang and Jianghao Lin},
      year={2026},
      eprint={2605.16986},
      archivePrefix={arXiv},
      primaryClass={cs.CL},
      url={https://arxiv.org/abs/2605.16986}, 
}

@misc{zhang2026mmskill,
      title={MMSkills: Towards Multimodal Skills for General Visual Agents}, 
      author={Kangning Zhang and Shuai Shao and Qingyao Li and Jianghao Lin and Lingyue Fu and Shijian Wang and Wenxiang Jiao and Yuan Lu and Weiwen Liu and Weinan Zhang and Yong Yu},
      year={2026},
      eprint={2605.13527},
      archivePrefix={arXiv},
      primaryClass={cs.AI},
      url={https://arxiv.org/abs/2605.13527}, 
}

@misc{Zhang2026SkillToLoRA,
      title={Skill-to-LoRA: From Using Skills to Learning Behaviors for Token-Efficient LLM Agents}, 
      author={Tianyi Zhang and Zhonghao Qi},
      year={2026},
      eprint={2606.16769},
      archivePrefix={arXiv},
      primaryClass={cs.AI},
      url={https://arxiv.org/abs/2606.16769}, 
}

@misc{Zhao2026ParametricSkills,
      title={Parametric Skills}, 
      author={Xuan Zhao and Haonan He and Qingyu Yang and Minglei Li and Jingqi Ye and Zelin Tan and Bo Wan and Peng Ye},
      year={2026},
      eprint={2606.30015},
      archivePrefix={arXiv},
      primaryClass={cs.CL},
      url={https://arxiv.org/abs/2606.30015}, 
}

@misc{yao2023webshopscalablerealworldweb,
      title={WebShop: Towards Scalable Real-World Web Interaction with Grounded Language Agents}, 
      author={Shunyu Yao and Howard Chen and John Yang and Karthik Narasimhan},
      year={2023},
      eprint={2207.01206},
      archivePrefix={arXiv},
      primaryClass={cs.CL},
      url={https://arxiv.org/abs/2207.01206}, 
}

@misc{zhang2026mmskillsmultimodalskillsgeneral,
      title={MMSkills: Towards Multimodal Skills for General Visual Agents}, 
      author={Kangning Zhang and Shuai Shao and Qingyao Li and Jianghao Lin and Lingyue Fu and Shijian Wang and Wenxiang Jiao and Yuan Lu and Weiwen Liu and Weinan Zhang and Yong Yu},
      year={2026},
      eprint={2605.13527},
      archivePrefix={arXiv},
      primaryClass={cs.AI},
      url={https://arxiv.org/abs/2605.13527}, 
}

\end{document}